\documentclass[11pt]{article}

\usepackage{PRIMEarxiv}

\usepackage[utf8]{inputenc} 
\usepackage[T1]{fontenc}    

\usepackage{hyperref,url}            
\usepackage{booktabs,multirow,multicol}       
\usepackage{amssymb,amsfonts,amsmath}       
\usepackage{nicefrac}   
\usepackage{enumitem}
\usepackage{microtype}      
\usepackage{lipsum}
\usepackage{fancyhdr}
\usepackage{graphicx,float}        

\usepackage[table]{xcolor}
\usepackage[sort,compress]{natbib}
\usepackage{multicol}

\title{T-SNN: Temporal Simplicial Neural Network \\for EEG Decoding}

\newcommand{\authoraffil}[1]{{\small\begin{tabular}[t]{@{}c@{}}#1\end{tabular}}}

\author{
  Nikita Malik\thanks{Equal contribution.} \\
  \authoraffil{Bharti School of Telecommunication\\ Technology and Management\\ Indian Institute of Technology, Delhi}
  \And
  Shubhajit Roy\footnotemark[1] \\
  \authoraffil{Department of Computer Science\\ \& Engineering\\ Indian Institute of Technology, Gandhinagar}
  \And
  Mohit Kataria\footnotemark[1] \\
  \authoraffil{Yardi School of\\ Artificial Intelligence\\ Indian Institute of Technology, Delhi}
  \AND
  Isuru Herath \\
  \authoraffil{Medical Scientist Training Program\\ School of Medicine\\ University of Pittsburgh}
  \And
  Suraj Yadav \\
  \authoraffil{Hertie Institute for AI\\ in Brain Health\\ University of T\"ubingen}
  \And
  In\'es Garc\'{\i}a-Redondo \\
  \authoraffil{AIDOS Lab\\ University of Fribourg}
  \AND
  Dhananjay Bhaskar\thanks{Correspondence: \texttt{dhananjay.bhaskar@wisc.edu}} \\
  \authoraffil{Department of Biomedical Engineering, Biophysics Graduate Program,\\
    Data Science Institute, Center for Genomic Science Innovation,\\
    Wisconsin Institute for Translational Neuroengineering\\
    University of Wisconsin--Madison}
}

\begin{document}
\maketitle
\setcounter{footnote}{0} 

\vspace{-.4cm}

\begin{abstract}
Decoding brain states requires models that capture both the evolution of neural activity and interactions among groups of brain regions. Existing EEG methods often treat recordings as multivariate time series or represent functional connectivity with pairwise graphs, leaving dynamic higher-order interactions largely unmodeled. We introduce the \emph{Temporal Simplicial Neural Network} (T-SNN), which represents EEG recordings as sequences of evolving simplicial complexes. By combining simplicial convolutions with recurrent updates, T-SNN jointly learns higher-order interactions and their temporal evolution. On the seven-class SEED-VII emotion recognition task, T-SNN outperforms convolutional, recurrent, graph-based, and Transformer methods in both trial-wise and cross-subject evaluations. Incorporating eye-movement features further improves performance, demonstrating the framework's potential for multimodal brain-state decoding.
\end{abstract}


\raggedcolumns 
\begin{multicols}{2}

\section{Introduction}
\label{sec:intro}

Inferring cognitive states, emotions, and sensory stimuli from neural activity is a central challenge in computational neuroscience and neuroAI. Electroencephalography (EEG) offers the temporal resolution needed to track rapid changes in brain activity, yet the information relevant to decoding may be distributed across recording sites and shift over the course of a trial. Pairwise connectivity alone may also miss interactions involving larger groups: higher-order coupling can encode synergistic information \cite{battiston2020networks,battiston2021physics,luppi2022synergistic,varley2023partial,schneidman2006weak}, reveal functional structures beyond conventional graph measures \cite{petri2014homological,reimann2017cliques}, improve brain-state decoding \cite{santoro2023higher}, and produce distinct dynamical regimes \cite{gambuzza2021stability,skardal2019abrupt,kim2022explosive}. 

Despite growing evidence that higher-order interactions contribute to brain dynamics, they remain largely absent from EEG decoding models. Convolutional, recurrent, graph-based, and Transformer methods have advanced emotion recognition \cite{li2018hcnn,zhong2022rgnn,vaswani2017,jiang2021gcnca,jiang2025maet}, but capture only parts of this structure. Sequence models learn temporal patterns while generally leaving interactions among recording sites implicit, graph-based models represent connectivity explicitly but usually restrict it to pairs. Moreover, changes in connectivity are often summarized across time or processed separately from spatial message passing, limiting the ability to model how group-level interactions reorganize during a trial.

To address this gap, we introduce the \emph{Temporal Simplicial Neural Network} (T-SNN), which jointly models higher-order 
\begin{figure*}[ht]
    \centering
    \includegraphics[width=\textwidth]{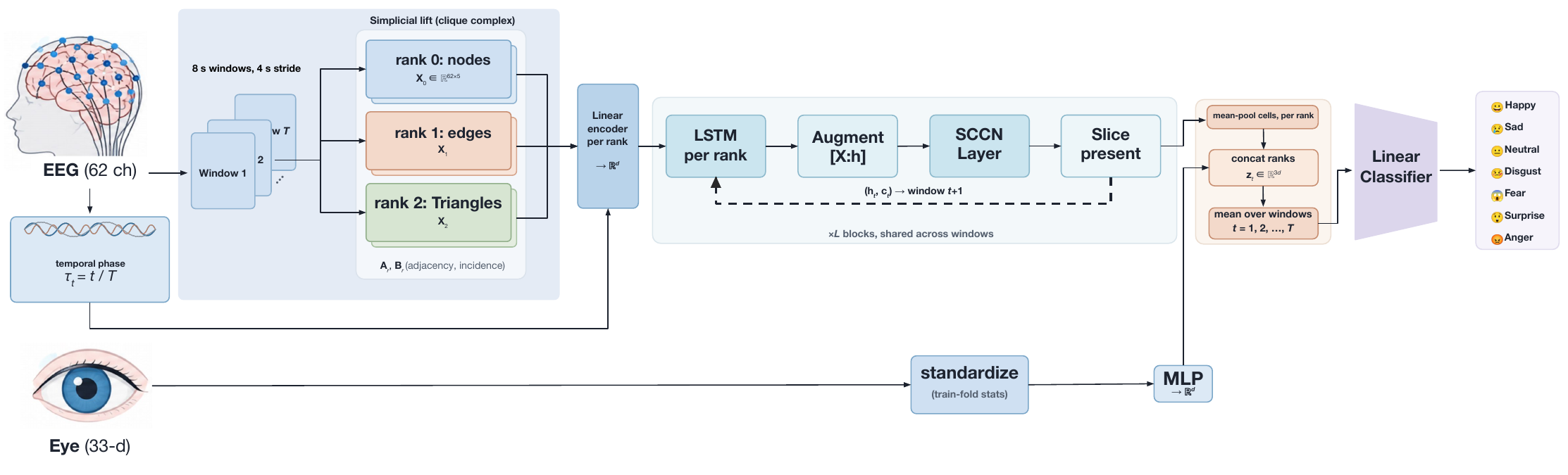}
    \caption{Overview of T-SNN. EEG windows are lifted into simplicial complexes and processed by blocks that jointly perform spatial and temporal message passing through simplicial convolutions and recurrent updates. The resulting representations are fused with eye-movement features for emotion recognition.
    \label{fig:schematic}
}
\end{figure*}
interactions and their temporal evolution through spatiotemporal message passing on simplicial complexes. T-SNN represents overlapping EEG windows as a sequence of evolving simplicial complexes, encoding individual channels as vertices, pairwise interactions as edges, and interactions among groups of channels as higher-order simplices. Simplicial convolutions exchange information within and across these structures, while recurrent states carry their history between windows. Information from other modalities can then be fused with the learned representation for multimodal decoding.

We evaluate T-SNN on seven-class emotion recognition using the SEED-VII dataset \cite{jiang2025maet}. T-SNN outperforms current methods in both accuracy and F1 score under trial-wise and leave-one-subject-out evaluations, achieving accuracies of $70.94\%$ and $69.87\%$, respectively. Multimodal decoding combining EEG and eye-movement features increases trial-wise accuracy to $77.50\%$. In summary, our main contributions are: (1) representation of dynamic EEG functional connectivity as time-varying simplicial complexes that explicitly encode higher-order interactions, (2) a recurrent simplicial architecture for joint spatial and temporal message passing, and (3) state-of-the-art performance on SEED-VII in trial-wise, cross-subject, and multimodal emotion recognition.

\section{Methods}
\label{sec:method}

We represent each EEG trial as a sequence of simplicial complexes. A trial is divided into $T$ overlapping windows $w_1,\dots,w_T$. Each window is lifted to a complex with cells of rank $r\in\{0,\dots,R\}$. A recurrent simplicial network passes messages within each window and carries information between windows. Here, $R$ is the maximum cell rank, $d$ is the shared hidden width, and $L$ is the number of stacked spatiotemporal blocks. We perform emotion classification using the learned spatiotemporal representations (Fig.~\ref{fig:schematic}).

\subsection{Simplicial Lift}
\label{subsec:lift}

For each EEG window $w_t$, we compute pairwise Pearson correlations $\rho_{ij}^{(t)}$ among channels. We retain pairs whose absolute correlation exceeds a threshold and use the resulting graph to construct a clique complex up to rank $R$. Each channel is a $0$-simplex, each retained pair is a $1$-simplex, and a set of $r+1$ channels forms an $r$-simplex when all its pairwise edges are present. We collect the rank-$r$ cell features in $\mathbf{X}_r^{(t)}\in\mathbb{R}^{n_r^{(t)}\times F}$, where $F$ is the number of frequency bands.

Node features are Welch log-band-power~\cite{welch1967use}. If
$\widehat{P}_i^{(t)}(f)$ is the power spectral density of channel $i$, its feature for band $b$ on $[f_b^-,f_b^+)$ is
\begin{equation}
  \bigl(\mathbf{x}_i^{(t)}\bigr)_b
  =
  \log\Bigl(
    \int_{f_b^-}^{f_b^+} \widehat{P}_i^{(t)}(f)\,df
    + \varepsilon
  \Bigr).
\end{equation}
For a simplex $s$ of rank $r\geq1$ in window $w_t$, we denote its weight by $\omega_s^{(t)}$. We set this weight to the mean correlation over all pairs of channels in $s$, then scale the mean of their node features by that weight:
\begin{equation}
  \omega_s^{(t)}
  =
  \frac{1}{\binom{r+1}{2}}
  \sum_{\substack{i,j\in s\\i<j}}\rho_{ij}^{(t)},
  \qquad
  \mathbf{x}_s^{(t)}
  =
  \frac{\omega_s^{(t)}}{r+1}\sum_{i\in s}\mathbf{x}_i^{(t)}.
\end{equation}
An edge therefore has its pairwise correlation as its weight, while a triangle has the mean weight of its three edges. The complex for window $w_t$ also defines a sparse same-rank adjacency $\mathbf{A}_r^{(t)}\in\mathbb{R}^{n_r^{(t)}\times n_r^{(t)}}$ and an incidence matrix
$\mathbf{B}_r^{(t)}\in\mathbb{R}^{n_{r-1}^{(t)}\times n_r^{(t)}}$ between ranks $r{-}1$ and $r$.

\vspace{-.2cm}
\subsection{Encoding and Temporal Phase}
A separate linear encoder is used for each rank to project cell
features to the shared width $d$: $\mathbf{H}_r^{(t)}=\mathbf{X}_r^{(t)}\mathbf{W}_r$, where
$\mathbf{W}_r\in\mathbb{R}^{F\times d}$. To encode when each window occurs within a trial, we normalize its position 
$\tau_t=t/T\in[0,1]$ and expand it on a geometric sinusoidal
\begin{table*}[!t]
\centering
\caption{Trial-wise SEED-VII accuracy and F1 score across EEG frequency bands (Avg.$\pm$Std., \%)}
\label{tab:seedvii_unimodal}
\small
\resizebox{2\columnwidth}{!}{
\begin{tabular}{llrrrrrr}
\toprule
& Method & Delta band & Theta band & Alpha band & Beta band & Gamma band & All bands\\
\midrule
\multirow{10}{*}{\rotatebox{90}{Accuracy}}
& KNN \cite{cover1967knn}           & $28.21\pm5.76$ & $28.47\pm7.02$ & $31.92\pm6.51$ & $34.83\pm5.77$ & $37.20\pm6.06$ & $36.43\pm5.38$ \\
& HCNN \cite{li2018hcnn}            & $42.33\pm6.36$ & $42.63\pm5.28$ & $43.62\pm4.82$ & $47.79\pm6.77$ & $48.18\pm6.10$ & $52.42\pm6.47$ \\
& RGNN \cite{zhong2022rgnn}         & $42.31\pm5.14$ & $41.73\pm5.53$ & $43.93\pm4.98$ & $44.68\pm5.51$ & $45.49\pm4.73$ & $48.50\pm6.83$ \\
& Transformer \cite{vaswani2017}    & $46.87\pm4.10$ & $47.01\pm4.23$ & $48.78\pm5.39$ & $53.10\pm6.34$ & $53.96\pm6.60$ & $56.04\pm7.82$ \\
& GCNCA \cite{jiang2021gcnca}       & $49.97\pm4.94$ & $50.46\pm4.29$ & $53.26\pm5.45$ & $58.36\pm6.69$ & $59.50\pm5.77$ & $58.04\pm7.78$ \\
& MAET \cite{jiang2025maet}         & --             & --             & --             & --             & --             & $58.11\pm8.78$ \\
& JODIE \cite{kumar2019jodie}       & $21.00\pm3.92$ & $21.81\pm5.20$ & $20.19\pm3.18$ & $18.25\pm3.86$ & $21.88\pm3.10$ & $23.06\pm4.88$ \\
& GraphMixer \cite{cong2023graphmixer} & $20.44\pm1.43$ & $19.31\pm1.65$ & $21.88\pm2.24$ & $18.56\pm1.80$ & $18.00\pm2.88$ & $24.88\pm1.01$ \\
& TGN \cite{rossi2020tgn}           & $22.75\pm1.87$ & $22.88\pm1.36$ & $23.06\pm1.89$ & $22.56\pm1.53$ & $23.44\pm2.19$ & $27.00\pm3.21$ \\
\cmidrule(l){2-8}
& \textbf{T-SNN} (ours)            & $\mathbf{70.50\pm2.72}$ & $\mathbf{73.00\pm1.59}$ & $\mathbf{69.94\pm1.39}$ & $\mathbf{66.25\pm1.49}$ & $\mathbf{66.44\pm2.18}$ & $\mathbf{70.94\pm1.86}$ \\
\midrule
\multirow{10}{*}{\rotatebox{90}{F1 score}}
& KNN \cite{cover1967knn}           & $26.00\pm5.98$ & $26.12\pm7.25$ & $29.67\pm6.64$ & $33.22\pm5.59$ & $34.98\pm5.62$ & $34.08\pm5.79$ \\
& HCNN \cite{li2018hcnn}            & $37.85\pm7.78$ & $38.30\pm5.72$ & $39.66\pm5.23$ & $44.05\pm7.55$ & $44.33\pm6.68$ & $49.02\pm6.80$ \\
& RGNN \cite{zhong2022rgnn}         & $37.85\pm5.56$ & $36.13\pm5.79$ & $39.18\pm5.43$ & $40.68\pm6.61$ & $41.77\pm5.50$ & $45.32\pm7.20$ \\
& Transformer \cite{vaswani2017}    & $43.20\pm4.26$ & $43.85\pm4.74$ & $45.48\pm5.23$ & $49.81\pm6.70$ & $51.48\pm6.96$ & $53.35\pm8.30$ \\
& GCNCA \cite{jiang2021gcnca}       & $47.61\pm5.14$ & $47.92\pm5.01$ & $50.99\pm5.71$ & $55.97\pm7.12$ & $57.54\pm5.90$ & $55.48\pm8.30$ \\
& MAET \cite{jiang2025maet}         & --             & --             & --             & --             & --             & $54.98\pm9.45$ \\
& JODIE \cite{kumar2019jodie}       & $14.47\pm3.92$ & $12.43\pm3.31$ & $14.26\pm2.04$ & $10.51\pm3.12$ & $13.24\pm2.64$ & $18.24\pm6.31$ \\
& GraphMixer \cite{cong2023graphmixer} & $15.67\pm2.03$ & $14.58\pm2.66$ & $14.69\pm4.00$ & $14.75\pm2.72$ & $13.31\pm1.40$ & $21.32\pm2.08$ \\
& TGN \cite{rossi2020tgn}           & $20.57\pm2.81$ & $19.11\pm1.55$ & $19.85\pm0.50$ & $18.55\pm1.64$ & $18.50\pm3.00$ & $24.79\pm3.83$ \\
\cmidrule(l){2-8}
& \textbf{T-SNN} (ours)            & $\mathbf{69.98\pm3.05}$ & $\mathbf{72.62\pm1.67}$ & $\mathbf{69.77\pm1.43}$ & $\mathbf{65.62\pm1.27}$ & $\mathbf{65.74\pm2.14}$ & $\mathbf{70.58\pm1.91}$ \\
\bottomrule
\end{tabular}}
\end{table*}
basis $\left[\tau_t,\ \sin(2^{k}\pi\tau_t),\
\cos(2^{k}\pi\tau_t)\right]_{k=0}^{K-1}$, where $K$ is the number of frequencies. We project the resulting temporal phase embedding to $\mathbb{R}^{d}$ and broadcast it to every cell at every rank in window $w_t$.

\vspace{-.2cm}
\subsection{Simplicial Message Passing}
\label{subsec:sccn}

We use a simplicial complex convolutional network (SCCN) layer to exchange information among cells within each window
\cite{yang2022sccn}. For each rank-$r$ cell, we aggregate messages from cells of the same rank, rank-$(r{-}1)$ cells on its boundary, and rank-$(r{+}1)$ cells that it bounds. We combine these messages before applying the update nonlinearity:
\begin{multline}
\mathbf{H}_r' = \sigma\!\Big(
\mathbf{A}_r\mathbf{H}_r\boldsymbol{\Theta}_r^{r\to r}
+\mathbf{B}_r^{\!\top}\mathbf{H}_{r-1}\boldsymbol{\Theta}_r^{r-1\to r}\\+\mathbf{B}_{r+1}\mathbf{H}_{r+1}\boldsymbol{\Theta}_{r}^{r+1\to r}\Big)
\label{eq:sccn}
\end{multline}
where $\boldsymbol{\Theta}\in\mathbb{R}^{d\times d}$ are learned weights for each message direction and $\sigma$ is the nonlinearity. We omit terms involving ranks outside $0,\dots,R$.

\vspace{-.1cm}
\subsection{Joint Spatiotemporal Message Passing}
\label{subsec:sist}

We use $L$ blocks to combine information within each EEG window with information carried from earlier windows. Each block contains an SCCN layer and an LSTM cell~\cite{hochreiter1997lstm} for each rank.
Adapting SiST-GNN~\cite{sistgnn}, we let the LSTM produce a summary $\mathbf{H}_r^{p}$ from its state after window $w_{t-1}$. We then process this summary together with the current features in a single SCCN pass.

To allow current cells to receive messages from both sets of
features, we stack $\mathbf{H}^{(t)}$ and $\mathbf{H}^{p}$ and augment the adjacency and incidence matrices from Eqn.~\ref{eq:sccn}:
\begin{equation}
\widetilde{\mathbf{A}}_r=
\begin{bmatrix}\mathbf{A}_r+\mathbf{I} & \mathbf{A}_r+\mathbf{I}\\[2pt]
\mathbf{0} & \mathbf{0}\end{bmatrix},
\quad
\widetilde{\mathbf{B}}_r=
\begin{bmatrix}\mathbf{B}_r & \mathbf{B}_r\\[2pt]
\mathbf{0} & \mathbf{0}\end{bmatrix}
\label{eq:augment}
\end{equation}
The upper rows route messages to current cells from current and past features at the same and adjacent ranks. The lower rows produce no updates and are discarded. We therefore retain only the updated current features:
\begin{equation}
\mathbf{H}_r^{(t,\ell+1)} =
\Big[\mathrm{SCCN}\big([\mathbf{H}^{(t,\ell)};\mathbf{H}^{p,\ell}],
\widetilde{\mathbf{A}},\widetilde{\mathbf{B}}\big)_r\Big]_{1:n_r^{(t)}}
\label{eq:block}
\end{equation}
where $\ell=0,\dots,L-1$ indexes the blocks, $\mathbf{H}_r^{(t,0)}$ is the encoded input from Section~2.2, and $\mathbf{H}_r^{(t,L)}$ is passed to the readout.
We share the block weights across windows, while the LSTM state carries information forward. Because simplices may disappear and reappear as connectivity changes, we track their recurrent states by simplex identity rather than row position. This allows a returning simplex to resume its previous state.

\begin{figure*}[t]
    \centering
    \includegraphics[width=\textwidth]{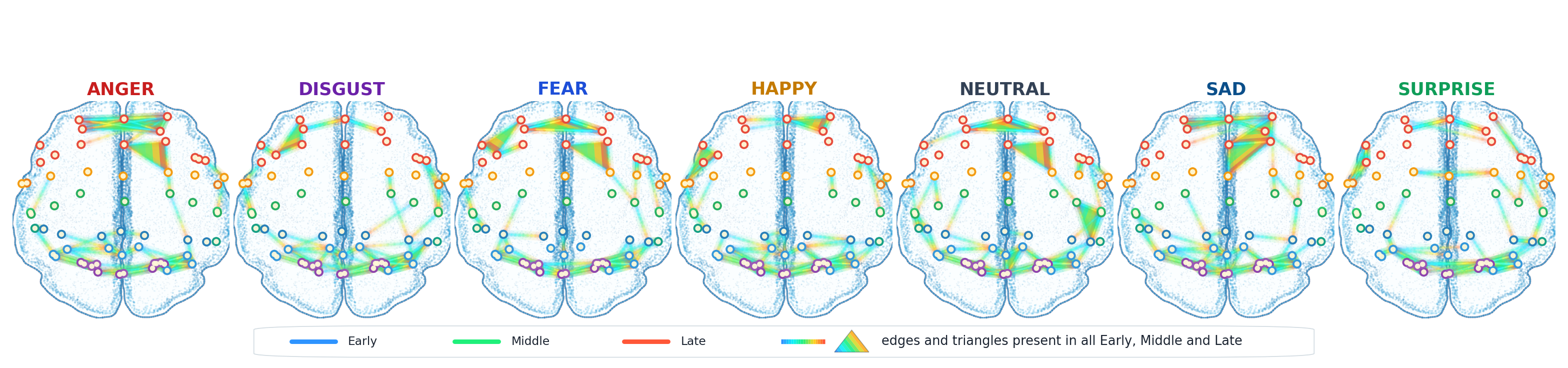}
    \caption{Group-averaged EEG simplices across seven emotions in the SEED-VII dataset ($N=20$). Lines represent edges (1-simplices), and filled triangles represent three-way interactions (2-simplices). Blue, green, and red indicate prevalence in early, middle, and late trial segments. Mixed colors indicate prevalence across multiple segments.}
    \label{visualisation_plot}
\end{figure*}

\subsection{Readout and Multimodal Fusion}
For each window, we mean-pool the cell features within each rank and concatenate the pooled features to obtain
$\mathbf{z}_t\in\mathbb{R}^{(R+1)d}$. When additional modalities are available, we standardize their features and use an MLP to map them to an embedding $\mathbf{e}_t\in\mathbb{R}^{d}$. We concatenate this
embedding

with $\mathbf{z}_t$, average across windows, and classify the trial:
\begin{table}[H]
\centering
\caption{Trial-wise SEED-VII accuracy and F1 score using multimodal EEG and eye movement data (Avg./Std., \%)}
\label{tab:seedvii_multimodal}
\resizebox{.9\columnwidth}{!}{
\begin{tabular}{lrrrr}
\toprule
\multirow{2}{*}{Method} & \multicolumn{2}{c}{Accuracy} & \multicolumn{2}{c}{F1 score} \\
\cmidrule(lr){2-3}\cmidrule(lr){4-5}
& Avg. & Std. & Avg. & Std. \\
\midrule
KNN \cite{cover1967knn}                & $40.44$ & $6.30$ & $37.94$ & $6.54$ \\
BDAE \cite{liu2016bdae}                & $61.55$ & $8.74$ & $59.11$ & $8.87$ \\
ETF \cite{wang2021etf}                 & $65.30$ & $8.55$ & $63.13$ & $8.88$ \\
VigilanceNet \cite{cheng2022vigilance} & $62.93$ & $7.12$ & $60.46$ & $7.81$ \\
MAET \cite{jiang2025maet}           & $71.28$ & $7.74$ & $69.16$ & $8.35$ \\
\midrule
\textbf{T-SNN} (EEG only)         & $70.94$ & $1.86$ & $70.58$ & $1.91$ \\
\textbf{T-SNN} (EEG + eye)        & $\mathbf{77.50}$ & $\mathbf{0.68}$ & $\mathbf{77.45}$ & $\mathbf{0.90}$ \\
\bottomrule
\end{tabular}}
\end{table}
\begin{equation}
\hat{\mathbf{y}}=\mathrm{softmax}\!\left(\mathbf{W}\,
\frac{1}{T}\sum_{t=1}^{T}\big[\mathbf{z}_t;\mathbf{e}_t\big]+\mathbf{b}\right)
\end{equation}
For EEG-only decoding, we omit $\mathbf{e}_t$.
\vspace{-.1cm}
\section{Implementation Details}
\label{sec:implementation}

\textbf{EEG features and simplicial construction.} We use $62$ EEG channels and divide each trial into 8-second windows with a 4-second stride. We retain pairs above the $97$th percentile of $|\rho_{ij}^{(t)}|$ within each window and construct a clique complex up to rank $R=2$. We set $\varepsilon=10^{-8}$ and use the Delta, Theta, Alpha, Beta and Gamma bands (1-4 Hz, 4-8 Hz, 8-14 Hz, 14-31 Hz, and 31-50 Hz respectively), giving $F=5$ for all bands or $F=1$ for a single band.

\noindent \textbf{Model configuration.} We use hidden width $d=512$, $L=2$ spatiotemporal blocks, and $K=8$ temporal phase frequencies. The SCCN update
(Eqn.~\ref{eq:sccn}) sums messages and uses a sigmoid for $\sigma$. We mean-pool cells by rank and average window embeddings across each trial.

\noindent \textbf{Training and model selection.} We train for $100$ epochs with Adam, learning rate $3\times10^{-4}$, cross-entropy loss, and gradient-norm clipping at $5.0$. We select the epoch with 
\begin{table}[H]
\centering
\caption{Leave-one-subject-out SEED-VII accuracy and F1 score (Avg./Std., \%)}
\label{tab:seedvii_crosssubject}
\resizebox{\columnwidth}{!}{
\begin{tabular}{lrrrr}
\toprule
\multirow{2}{*}{Method} & \multicolumn{2}{c}{Accuracy} & \multicolumn{2}{c}{F1 score} \\
\cmidrule(lr){2-3}\cmidrule(lr){4-5}
& Avg. & Std. & Avg. & Std. \\
\midrule
KNN \cite{cover1967knn}          & $20.85$ & $4.56$ & $20.23$ & $4.49$ \\
HCNN \cite{li2018hcnn}           & $39.88$ & $4.94$ & $38.18$ & $5.06$ \\
RGNN \cite{zhong2022rgnn}        & $37.49$ & $5.44$ & $34.52$ & $4.83$ \\
Transformer \cite{vaswani2017}   & $40.36$ & $5.22$ & $37.76$ & $5.54$ \\
GCNCA \cite{jiang2021gcnca}      & $38.68$ & $\mathbf{3.94}$ & $37.25$ & $\mathbf{3.65}$ \\
CLISA \cite{shen2023clisa}       & $38.27$ & $5.23$ & $34.02$ & $5.34$ \\
MAET (w/o AT) \cite{jiang2025maet} & $40.69$ & $5.50$ & $38.47$ & $6.09$ \\
MAET \cite{jiang2025maet}     & $40.90$ & $5.52$ & $38.85$ & $6.07$ \\
\midrule
\textbf{T-SNN}                    & $\mathbf{69.87}$ & $6.49$ & $\mathbf{68.26}$ & $7.55$ \\
\bottomrule
\end{tabular}}
\end{table}
the highest validation macro-F1 score. We assemble trials into block-diagonal minibatches for efficiency using sparse message-passing operations, using batch sizes of
$64$ for trial-wise and $256$ for cross-subject evaluation. 

\noindent Our code is available on GitHub\footnote{\url{https://github.com/UW-Madison-CBML/temporal-simplicial-nn}}
\vspace{-.4cm}
\section{Results}

We evaluate seven-class emotion recognition on SEED-VII \cite{jiang2025maet}, which
contains $80$ trials from each of $20$ subjects. We compare T-SNN with KNN
\cite{cover1967knn}, HCNN \cite{li2018hcnn}, RGNN \cite{zhong2022rgnn}, GCNCA \cite{jiang2021gcnca}, and CLISA \cite{shen2023clisa}, a Transformer \cite{vaswani2017}, MAET \cite{jiang2025maet},
the temporal graph models JODIE \cite{kumar2019jodie}, GraphMixer \cite{cong2023graphmixer}, and TGN
\cite{rossi2020tgn}, and the
multimodal methods BDAE \cite{liu2016bdae}, ETF \cite{wang2021etf}, and VigilanceNet \cite{cheng2022vigilance}.

We evaluate both trial-wise and cross-subject decoding. For the trial-wise setting, we use four-fold cross-validation over the $1600$ subject--trial pairs, stratified by emotion. For the cross-subject setting, we hold out one subject per fold. We select validation data from the training folds and compute eye-movement
standardization statistics using training data only. We report mean accuracy and F1 score with their standard deviations across folds. Fig.~\ref{visualisation_plot} shows the group-averaged edges and triangles for each emotion and their prevalence during early, middle, and late portions of the trials.

For EEG-only trial-wise decoding, T-SNN achieves the highest accuracy and F1 score across every frequency-band setting in Table~\ref{tab:seedvii_unimodal}. Its strongest single-band results occur in theta, whereas the strongest baseline results occur in beta or gamma. JODIE, GraphMixer, and TGN reach at most
$27\%$ accuracy, suggesting that models designed for streams of edge events are less suited to the window-based connectivity used here. Adding eye-movement features improves trial-wise accuracy from $70.94\%$ to $77.50\%$ and F1 score from $70.58\%$ to $77.45\%$ (Table~\ref{tab:seedvii_multimodal}). The accuracy standard deviation also decreases from $1.86$ to $0.68$.

In leave-one-subject-out evaluation, T-SNN reaches $69.87\%$ accuracy and a $68.26\%$ F1 score, exceeding all methods in Table~\ref{tab:seedvii_crosssubject}. Its accuracy standard deviation of $6.49$ reflects variation across held-out subjects. On an NVIDIA H200 GPU, T-SNN takes about $50$\,s per epoch, compared with $128$\,s, $354$\,s, and $439$\,s for JODIE, GraphMixer, and TGN. These runs make T-SNN $2.5$--$9\times$ faster despite its larger hidden width ($d=512$ versus $d=100$).

\vspace{-.1cm}
\section{Conclusions}

We introduced T-SNN, a temporal simplicial neural network for EEG decoding that models evolving higher-order interactions among brain regions. On the SEED-VII emotion recognition task, T-SNN outperformed the evaluated methods in both trial-wise and cross-subject settings, while incorporating eye-movement features further improved performance. These results demonstrate the value of capturing higher-order interactions for complex decoding tasks. Future work will focus on interpreting the inferred structures in neurophysiological terms and exploring alternative ways to construct simplices, including approaches based on Granger causality. Beyond brain activity, T-SNN offers a general framework for modeling time-varying higher-order interactions in molecular and cellular systems and in population dynamics.

\vspace{-.15cm}
\section*{Acknowledgments}
\vspace{-.2cm}
This work was supported in part by the IEEE Signal Processing Society (SPS) SigMA program.

\end{multicols}

\bibliographystyle{unsrt}  
\bibliography{refs}  

\end{document}